\documentclass[journal]{IEEEtran}

\usepackage{cite}
\ifCLASSINFOpdf \usepackage[pdftex]{graphicx} \else \usepackage[dvips]{graphicx} \fi

\ifCLASSOPTIONcompsoc \usepackage[caption=false,font=normalsize,labelfont=sf,textfont=sf]{subfig} \else \usepackage[caption=false,font=footnotesize]{subfig} \fi \usepackage{amsmath} 

\usepackage{amssymb}
\usepackage{tabularx}

\DeclareMathOperator*{\argmin}{arg\,min}
\newcolumntype{P}[1]{>{\centering\arraybackslash}p{#1}}

\begin{document}

\title{Accurate Ground-Truth Depth Image Generation via Overfit Training of Point Cloud Registration using Local Frame Sets}

\author{Jiwan~Kim,
        Minchang~Kim,
        Yeong-Gil Shin,
        and~Minyoung~Chung$^{\ast}$

\thanks{\textit{Asterisk indicates corresponding author.\\(This paper is under consideration at Computer Vision and Image Understanding.)}}
\thanks{J. Kim, M. Kim, and Y.-G. Shin are with the Department of Computer Science and Engineering, Seoul National University, South Korea.}
\thanks{*M. Chung is with the School of Software, Soongsil University, South Korea (e-mail: chungmy@ssu.ac.kr).}}

% \markboth{This paper is under consideration at Computer Vision and Image Understanding.}
% {Shell \MakeLowercase{\textit{et al.}}: Accurate Depth Ground-Truth Generation via Overfit-Training of Point Cloud Registration using Local Frame Sets}

\maketitle

\begin{abstract}
   Accurate three-dimensional perception is a fundamental task in several computer vision applications. Recently, commercial RGB-depth (RGB-D) cameras have been widely adopted as single-view depth-sensing devices owing to their efficient depth-sensing abilities. However, the depth quality of most RGB-D sensors remains insufficient owing to the inherent noise from a single-view environment. Recently, several studies have focused on the single-view depth enhancement of RGB-D cameras. Recent research has proposed deep-learning-based approaches that typically train networks using high-quality supervised depth datasets. The performance of current learning-based depth enhancement techniques is highly dependent on the quality of the ground-truth (GT) depth dataset; however, such high-quality GT datasets are difficult to obtain. In this study, we developed a novel method for high-quality GT depth generation based on an RGB-D stream dataset. First, we defined consecutive depth frames in a local spatial region as a local frame set. Then, the depth frames were aligned to a certain frame in the local frame set using an unsupervised point cloud registration scheme. The registration parameters were trained based on an overfit-training scheme, which was primarily used to construct a single GT depth image for each frame set. The final GT depth dataset was constructed using several local frame sets, and each local frame set was trained independently. The primary advantage of this study is that a high-quality GT depth dataset can be constructed under various scanning environments using only the RGB-D stream dataset, which can be used as a single-view depth supervision dataset. Moreover, our proposed method can be used as a new benchmark GT dataset for accurate performance evaluations. We evaluated our GT dataset on previously benchmarked GT depth datasets and demonstrated that our method is superior to state-of-the-art depth enhancement frameworks.

\end{abstract}


\begin{figure}[t]
    \begin{center}
    \includegraphics[width=\linewidth]{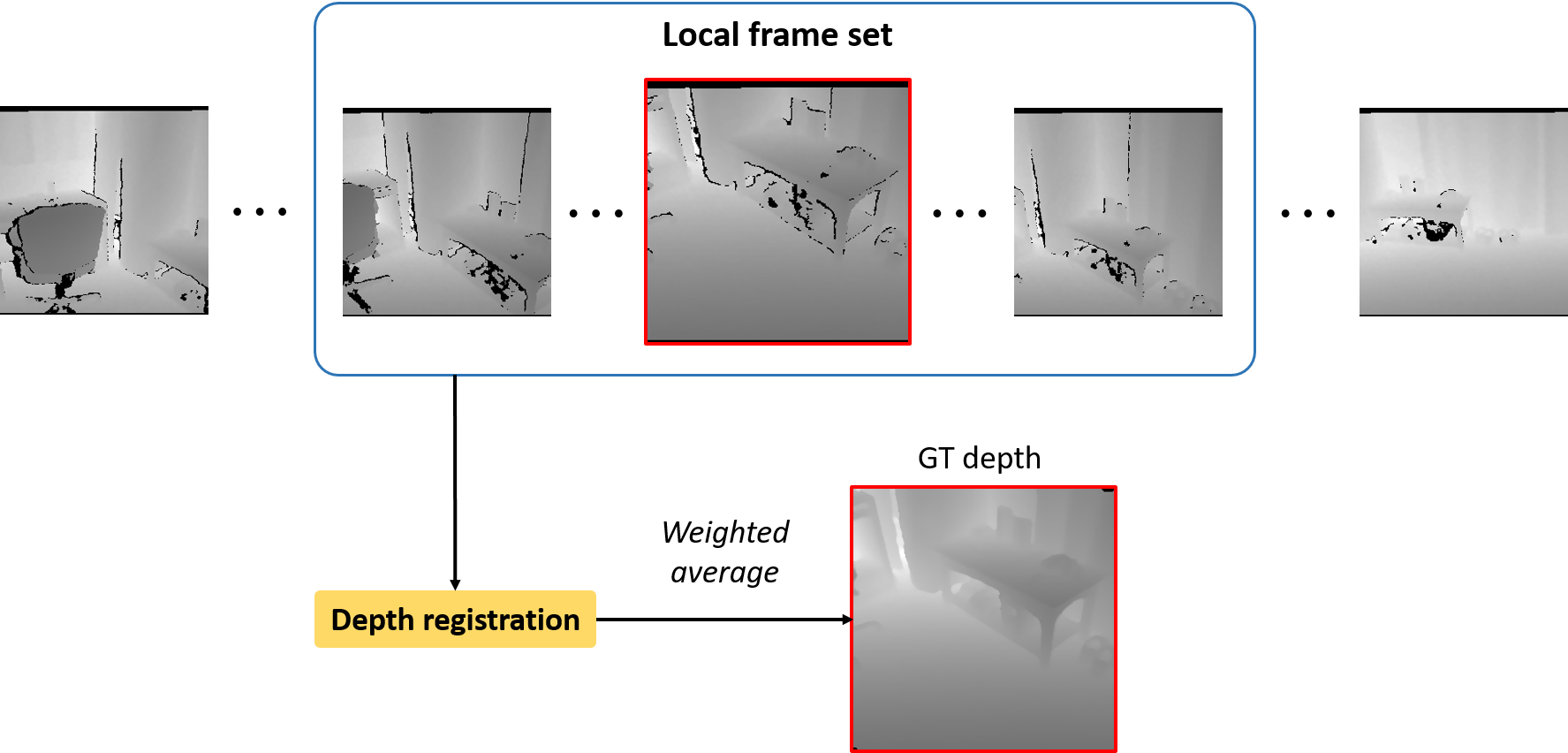}
    \end{center}
    \caption{Multiview-based ground-truth(GT) depth image generation. The registration parameters in the local frame sets are aligned to their local target frame (i.e., the box frame outlined in red in the local frame set). The weighted average of the aligned depth images in the frame set at the pixel-level is obtained to generate the GT depth image.}
    \label{fig:gt_gen}
\end{figure}

\section{Introduction}

Accurate depth perception is a prerequisite for several computer vision and robotics applications, such as simultaneous localization and mapping \cite{henry2012rgb, dai2017bundlefusion, schops2019bad}, object recognition \cite{eitel2015multimodal, schwarz2015rgb, aldoma2012global}, and object semantic segmentation \cite{mian2006three, dai2017scannet, tian2019objectfusion}. Recently, commercial RGB-D cameras (e.g., Kinect, Realsense, ASUS Xtion) have been widely adopted as single-view depth sensors owing to their affordable price and portability. However, they still suffer from insufficient depth quality due to heavy noise and missing values. Because deficient depth information can severely mislead downstreaming tasks, enhancing the depth quality when using a commercial depth camera is a fundamental task for achieving superior performance in three-dimensional (3D) vision applications. Several previous studies have performed single-view depth enhancement tasks based on traditional filter methods \cite{diebel2005application, ferstl2013image, lu2014depth, shen2015mutual}, and modern deep learning methods \cite{yan2018ddrnet, sterzentsenko2019self, jeon2018reconstruction, zhang2018deep}. Deep-learning-based methods show promising results over traditional approaches, which typically require a high-quality ground-truth (GT) depth dataset for training the networks. 
\par

Studies on depth enhancement strategies can be categorized based on the use of classical image processing methods or deep-learning-based methods. Traditional image processing approaches focus on high-quality RGB images that are synchronized with depth images \cite{diebel2005application, ferstl2013image, lu2014depth, shen2015mutual}. These methods leverage the abundant color texture information for guidance in recovering the depth image by modeling the correlation between the color and depth geometries. Because filter-based methods are vulnerable to heavy noise and missing values, deep-learning-based approaches have been proposed to address these issues. Several studies have attempted to model the noise that is inherent to raw input depth data \cite{johns2016deep, planche2017depthsynth, sweeney2019supervised}. As realistic noise is combined with various factors (e.g., light sources, materials, and distances), estimating a realistic noise model is difficult \cite{gu2020coupled}. Other effective approaches have been investigated to generate a reliable synthetic dataset that can be obtained using generative models \cite{shrivastava2017learning, bulat2018learn, gu2020coupled}. Owing to the difficulty in obtaining abundant real-world datasets, such approaches have been used to generate reliable synthetic GT datasets with realistic simulators \cite{hui2016depth, song2016deep}. Such synthetic-dataset-based methods require accurate scenes from real-world datasets \cite{gu2020coupled}. A few studies attempted to use a real-world dataset for supervision by incorporating their own scanning system for the task. These methods used multiview depth supervision as nonrigid reconstruction \cite{yan2018ddrnet}, and multicamera setting \cite{sterzentsenko2019self} for real-world depth-supervised approaches. Although the methods demonstrated superior results when compared to previous methods, such scanning systems have difficulty in constructing real-world datasets because they require fixed scanning environments. Consequently, an applicable real-world GT depth dataset is required.
\par

A real-world GT depth dataset is also required in the recently proposed depth enhancement applications using deep-learning-based approaches. The learning-based approaches show superior results by employing supervised metrics using the GT dataset. However, current methods still suffer from the real-world GT dataset collection problem, which has not yet been significantly explored. One possible approach for improving the GT depth generation process is to utilize frames with multiple views. Depth information from other views can be used as a supplement for missing areas in the single-view raw depth information. In this case, an accurate estimation of pose parameters (i.e., spatial registration) is critical for aligning multiple depth images from different views. In the last decade, considerable research has been conducted to construct real-world pose datasets that are obtained using RGB-D cameras \cite{firman-cvprw-2016}. These datasets provide RGB-D scanned images and a visual odometry dataset using a 3D reconstruction method \cite{meister2012can, dai2017scannet}, which indicates that the estimated camera poses were optimized with a global frame set. Inspired by these works, large-scale RGB-D and pose dataset-based approaches have been proposed \cite{jeon2018reconstruction, zhang2018deep}. These methods privileged the real-world 3D reconstruction dataset \cite{dai2017scannet, Matterport3D}, which were generated by projecting reconstructed meshes using the given poses. Such a dataset has been used in considerably novel works as a direct supervision dataset \cite{jeon2018reconstruction, zhang2018deep}, or for performance evaluations \cite{gu2020coupled}. However, such pose parameters have been estimated using classical handcrafted features; consequently, the dataset is relatively vulnerable to texture-less and noisy regions when compared to current datasets with deep-learning-based features \cite{zeng20173dmatch}. 
\par

To address these limitations, we present a novel method for generating a real-world GT dataset (Fig. \ref{fig:gt_gen}). Our method only requires an adequate number of neighboring frames for the GT generation of a certain depth frame, without the requirement of a GT pose dataset. First, we defined consecutive frames in a local spatial region as a local frame set consisting of a target frame and neighboring frames. Then, the depth frames in the local frame set were aligned to the target frame. A novel unsupervised point cloud registration scheme was adopted to estimate the relative camera pose parameters of the frames \cite{el2021unsupervisedr}. Our primary objective was to precisely estimate the aligned pose parameters optimized in the local frame set to reduce misalignment errors. To achieve this, the registration parameters were trained based on an unsupervised, overfit-training scheme. Then, the GT depth frame of a local set was generated by averaging the aligned frames to obtain a clean and dense depth dataset.
\par

The novelty of this works can be summarized as follows:

\begin{itemize}
    \item We propose an applicable method for a real-world GT depth dataset that does not require additional supervision.
    \item We generated a high-quality real-world depth dataset to improve the accuracy of the GT depth dataset using an overfit-trained unsupervised point cloud registration scheme.
    \item We propose a self-supervised depth image enhancement framework using only RGB-D stream dataset.
    \item We introduce a new benchmark GT depth dataset for accurate performance evaluation in future studies.
\end{itemize}

The remainder of this manuscript is structured as follows. An overview of related works is presented in Section 2. Section 3 describes the details of the proposed method. Section 4 presents the experimental results of this study. The discussion and conclusions of this study are presented in Sections 5 and 6, respectively.
\par

\begin{figure*}[t]
    \begin{center}
    \includegraphics[width=\linewidth]{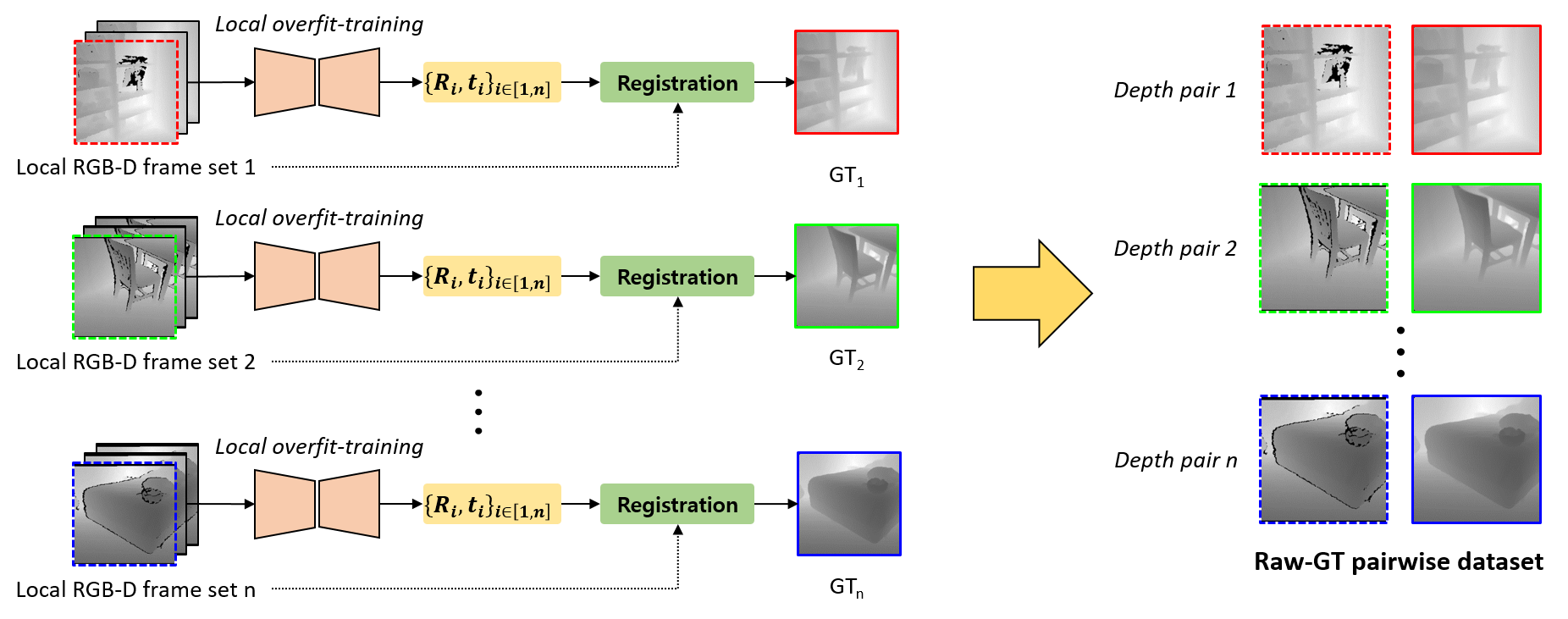}
    \end{center}
    \caption{Process of GT depth dataset generation. The depth pairs are overfit-trained individually in their own local frame set. $\textbf{\textit{R}}_{\textit{i}}$, $\textbf{\textit{t}}_{\textit{i}}$ represent the rotation matrix and translation vector of the $i^{th}$ depth frame in the corresponding local frame sets, respectively. The images outlined with colored (i.e., red, green, and blue) dashed lines are the original input depth images and the corresponding images outlined with colored solid lines are the generated GT depth images.}
    \label{fig:dataset_gen}
\end{figure*}

\section{Related Work}
The proposed method generates a GT depth image by averaging the aligned local frame set. To obtain the relative pose parameters of the frames, the transformation matrices of the frames were estimated using a point cloud registration scheme. Let $\{\textbf{P},\textbf{Q}\}\in \mathbb{R}^3$ be two point clouds from different frames of view. To align the point cloud $\textbf{Q}$ to $\textbf{P}$, the estimating optimal transformation matrix $\textbf{T}^*$ can be formulated as $\textbf{T}^* = \argmin_{\textup{T}}{\|\textbf{P}-\textup{T}(\textbf{Q})\|}$, which minimizes the distance between the point cloud $\textbf{P}$ and the transformed $\textbf{Q}$ (i.e., $\textup{T}(\textbf{Q})$). The general pipeline for point cloud registration consists of three main steps: feature descriptor extraction, matching correspondence, and transformation parameter estimation. In the following section, we briefly summarize the phases of the point cloud registration task. 
\par

\subsection{Classical Point Cloud Registration}
The traditional point cloud registration methods rely heavily on handcrafted feature descriptors. These methods estimate the relative poses directly from manually defined feature descriptors to determine geometric correspondence. In recent decades, several descriptors have been proposed to define geometric features using local 3D neighboring points, such as spin images \cite{johnson1999using}, geometry histograms \cite{frome2004recognizing}, and feature histograms \cite{rusu2008aligning, rusu2009fast}. Despite the improvements achieved by these approaches, their performance is still sensitive to the quality of data (e.g., noise, low resolution, missing values); moreover, these methods exhibit limitations in distinguishing the features in certain texture-less primitives, such as planes or smooth surfaces.
\par

\subsection{Learning-based Point Cloud Registration}

\subsubsection{Supervised Approach}
Recent research has proposed supervised learning methods to address the aforementioned problems by leveraging a pose GT dataset. These approaches attempt to improve the distinguishing ability by extracting deep-learning feature descriptors \cite{choy2019fully, gojcic2019perfect, yew20183dfeat, deng20193d, li2020end}, or determining accurate correspondence, which is directly used for the final parameter estimation step \cite{wang2019deep, yew2020rpm, choy2020deep, sarlin2020superglue}. These methods train high-level features from the surface dataset or highly consistent features from the given pose dataset. However, obtaining an accurate GT dataset is difficult, and the pretrained GT dataset may be biased toward its own dataset.
\par

\subsubsection{Unsupervised Approach} 
Several unsupervised approaches have been proposed recently to address the GT collection problem. These methods attempt to achieve the task based on various learning strategies, such as feature extraction \cite{li2019usip, kadam2021r}, geometric transformation \cite{wang2020unsupervised, huang2020feature}, and sampling distribution \cite{jiang2021sampling}. However, their aim is to perform registration in a sparse-object scale; thus, the application of these methods to the dense point cloud obtained from the RGB-D camera is time consuming. Recently, an unsupervised method for point cloud registration of data from the RGB-D dataset was proposed \cite{el2021unsupervisedr}. The method leverages differentiable alignment and rendering schemes to enforce unsupervised losses. This method enables dense point cloud registration from arbitrarily scanned RGB-D frames in a fully end-to-end unsupervised manner. Inspired by this work, we invented a multiview-based GT depth generation scheme that can be constructed using only an RGB-D stream dataset, without any other GT datasets. 
\par

\section{Methodology}

\subsection{Problem Formulation}

The pose optimization problem for all multiview frames can be formulated as follows:

\begin{equation}
    \begin{split}
        &\textbf{T}^*=\argmin_{\textbf{T}}{\sum_{i=1}^N\sum_{j=1}^N\|\textbf{P}_i-\textup{T}_{j \rightarrow i}(\textbf{P}_j)\|},\\
        &where \;i \neq j, \; \textbf{T}=\{\textup{T}_i(\cdot)\},
     \end{split}
    \label{eq:1}
\end{equation}
where $\textup{T}_i$ is the transformation matrix of frame $i$ to global target frame (i.e., world coordinate), and $\textup{T}_{j\rightarrow i}(\textbf{P}_j)$ indicates the transformation of point cloud $\textbf{P}_j$ to $\textbf{P}_i$. The optimal transformation matrix set $\textbf{T}^*$ is estimated for all $N$ scenes in the global target frame. However, the reconstructed meshes (i.e., 3D surface data) from the globally optimized pose parameters contain occasional misalignment \cite{gu2020coupled} and over-smoothing errors \cite{sterzentsenko2019self}, which can mislead the results of the deep-learning-based approaches. To address these problems, we propose a local-frame-set-based method to generate the GT depth image using depth frames in independent local spatial regions (\ref{fig:dataset_gen}). In this case, the pose parameters are estimated in each local frame set. For a point cloud $\textbf{P}_i$ from $i^{th}$ depth frame and its $n$ neighbor point cloud set $\textbf{P}_{j\in [1,n]}$, the point cloud registration problem of the neighboring frames to the local target frame (i.e., $i^{th}$ frame) can be represented by the sub-formulation of (\ref{eq:1}) as follows:
 
 \begin{equation}
    \begin{split}
      &\textbf{T}_i^*=\argmin_{\textbf{T}_i}{\sum_{j=1}^n\|\textbf{P}_i-\textup{T}_{j\rightarrow i}(\textbf{P}_j)\|},\\
      &where \; \textbf{T}_i=\{\textup{T}_{j\rightarrow i}(\cdot)\}.
     \end{split}
      \label{eq:2}
 \end{equation}
The transformation matrices $\textbf{T}_i^*$ of the frame sets are optimized independently, and each frame set is aligned to its local target frame, unlike the pose estimation of the entire frame as shown in (\ref{eq:1}). 
\par

\subsection{Depth Dataset Generation}
As illustrated in Fig. \ref{fig:gt_gen}, the proposed GT depth image generation method consists of two steps: local frame set alignment and depth rendering. The first step is achieved by the unsupervised point cloud registration scheme \cite{el2021unsupervisedr}, which is performed using the overfit-trained parameters. The subsequent rendering step is attained by projecting the aligned point clouds onto the local target frame using a pixel-level weighted averaging scheme. This process is performed in each local frame set, and the final GT depth dataset is constructed using several local frame sets, as illustrated in Fig. \ref{fig:dataset_gen}.
\par

\subsubsection{Unsupervised Point Cloud Registration}
The alignment of the local frame set is performed by estimating the relative pose parameters. To estimate the pose parameters between frames of different depths, a state-of-the-art point cloud registration scheme was adopted \cite{el2021unsupervisedr}. The authors used differentiable alignment and rendering strategy \cite{wiles2020synsin} to impose consistency losses between projected rendered point cloud and input image. With a set of $k$ corresponding points $\mathcal{M}=\{(t,s,\omega)_i : 0 \leq i < k\}$, three losses, i.e., depth, photometric, and correspondence losses are defined as follows:

\begin{figure*}[t]
    \begin{center}
    \includegraphics[width=\linewidth]{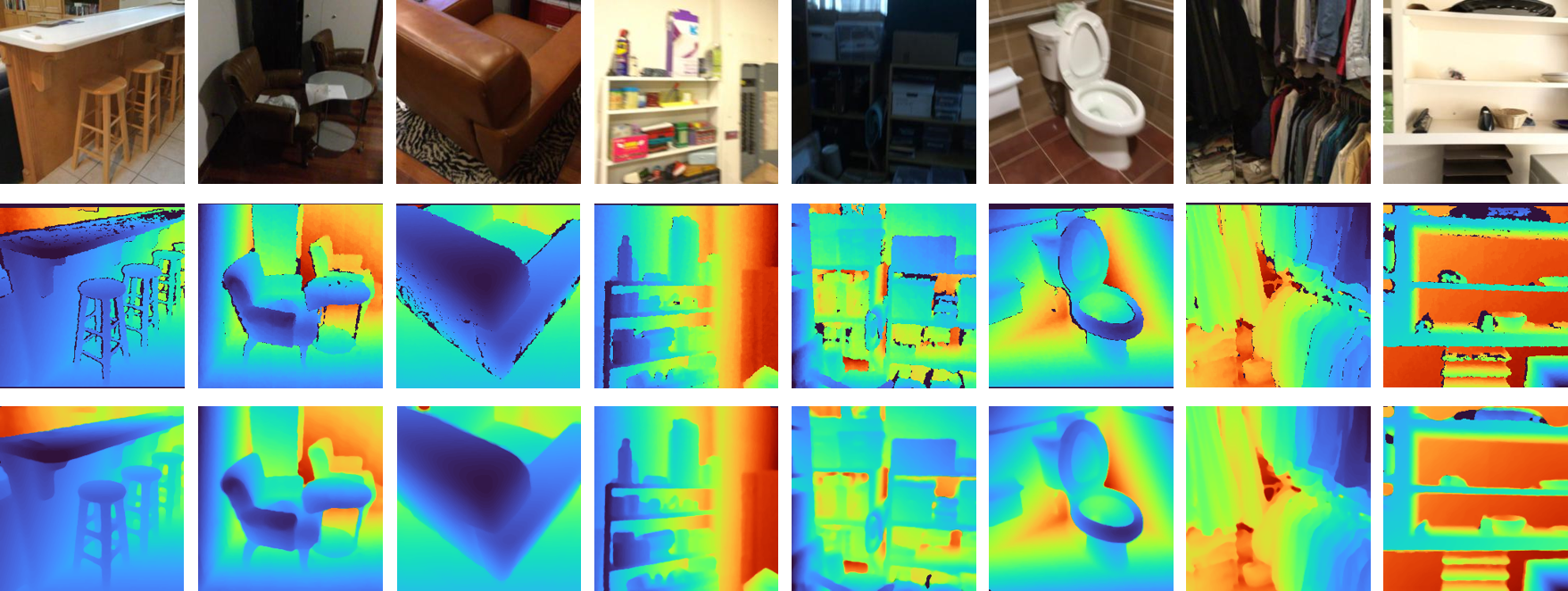}
    \end{center}
    \caption{GT depth generation results. Top row: RGB images; second row: input depth images; bottom row: GT images generated using the proposed method. Each column has corresponding frame.}
    \label{fig:our_gt}
\end{figure*}

\begin{equation}
    \begin{split}
        &\mathcal{L}_{D} = \sum_{\Omega(\textbf{d}_t)}{|\textbf{d}_t-Proj(\textbf{p}^{\textbf{x}}_{s \rightarrow t})|},\\
        &\mathcal{L}_{P} = \sum_{\Omega(\textbf{I}_t)}{|\textbf{I}_t-Proj(\textbf{p}^{rgb}_{s \rightarrow t})|},\\
        &\mathcal{L}_{C} = |\mathcal{M}|^{-1}\sum_{\mathcal{M}}{\omega(\textbf{p}^{\textbf{x}}_t-\textbf{p}^{\textbf{x}}_{s \rightarrow t})^2},\\
        &where \;\; \textbf{p}_{s \rightarrow t} = \textup{T}(\textbf{p}_s),
    \end{split}
    \label{eq:3}
\end{equation}
where $\textbf{p}=(\textbf{p}^{\textbf{x}}, \textbf{p}^{rgb})\in \mathbb{R}^6$ is a point which contains $\textbf{p}^{\textbf{x}}$, which is a 3D coordinate, and $\textbf{p}^{rgb}$, which indicates the color space. $\textbf{d} \in \Omega(\textbf{d})$ and $\textbf{I} \in \Omega(\textbf{I})$ indicate the depth, and RGB pixel value, respectively, $t$ and $s$ denote the elements of the target and source frame, respectively. $Proj(\textbf{p})$ denotes the projected rendered image from \textbf{p} according to its superscript. The distance ratio between the first and second nearest neighbor points from a source to the target is defined as $r = (D(t,s_{t,1})/D(t,s_{t,2}))$ and the corresponding weight between the 3D point $\textbf{x}_t$ and $\textbf{x}_s$ can be defined as $\omega = 1-r$. The consistency losses train the feature encoder to generate a unique correspondence between the two frames to derive the relative camera poses using input RGB-D frames. Contrast to existing pose-supervised point cloud registration approaches \cite{gojcic2020learning, detone2018superpoint, choy2019fully, choy2020deep}, this method is performed in a fully end-to-end unsupervised manner. This method can be applied to any other unannotated RGB-D stream dataset, which can be used for accurate GT depth generation. 
\par

\subsubsection{Local Frame Set Registration}
To attain a precisely aligned local frame set, the pose estimation problem in (\ref{eq:2}) was modified by employing the unsupervised learning method \cite{el2021unsupervisedr} for each local RGB-D frame set. Let us consider $k$ neighboring frames of the $i^{th}$ target frame. Then the loss functions of the local set can be represented by a summation of (\ref{eq:3}) as follows:

\begin{equation}
    \begin{split}
        &\mathcal{L}_{D_i} = \sum_{\Omega(\textbf{d}_t)}\sum_{j=1}^k{|\textbf{d}_t-Proj(\textbf{p}^{\textbf{x}}_{s,j \rightarrow t})|},\\
        &\mathcal{L}_{P_i} = \sum_{\Omega(\textbf{I}_t)}\sum_{j=1}^k{|\textbf{I}_t-Proj(\textbf{p}^{rgb}_{s,j\rightarrow t})|},\\
        &\mathcal{L}_{C_i} = \sum_{\mathbf{M}_i}\sum_{j=1}^k{\omega_j|\mathcal{M}_{i,j}|^{-1} (\textbf{p}^{\textbf{x}}_t-\textbf{p}^{\textbf{x}}_{s,j \rightarrow t})^2},
    \end{split}
    \label{eq:4}
\end{equation}
where $ \mathbf{M}_i = [\mathcal{M}_{i,1}, \mathcal{M}_{i,k}]$. For consistency between the target frame and the frames from the other rendered neighboring frames, the frames are trained simultaneously to precisely align every possible pair in the local frame set (i.e., alignment between a neighbor frame and the target frame, and between a neighbor frame and another neighbor frame). To derive robust corresponding points for accurate pose estimation in the frames, the features are trained only in the frame set, which implies that the features are overfit trained in a certain local frame set. The overfit-trained feature encoder yields feasible feature descriptors for the local frame set, and precise pose parameters are achieved by the corresponding coordinate geometry from the overfit-trained features. 
\par

\begin{figure}[t]
    \begin{center}
    \includegraphics[width=\linewidth]{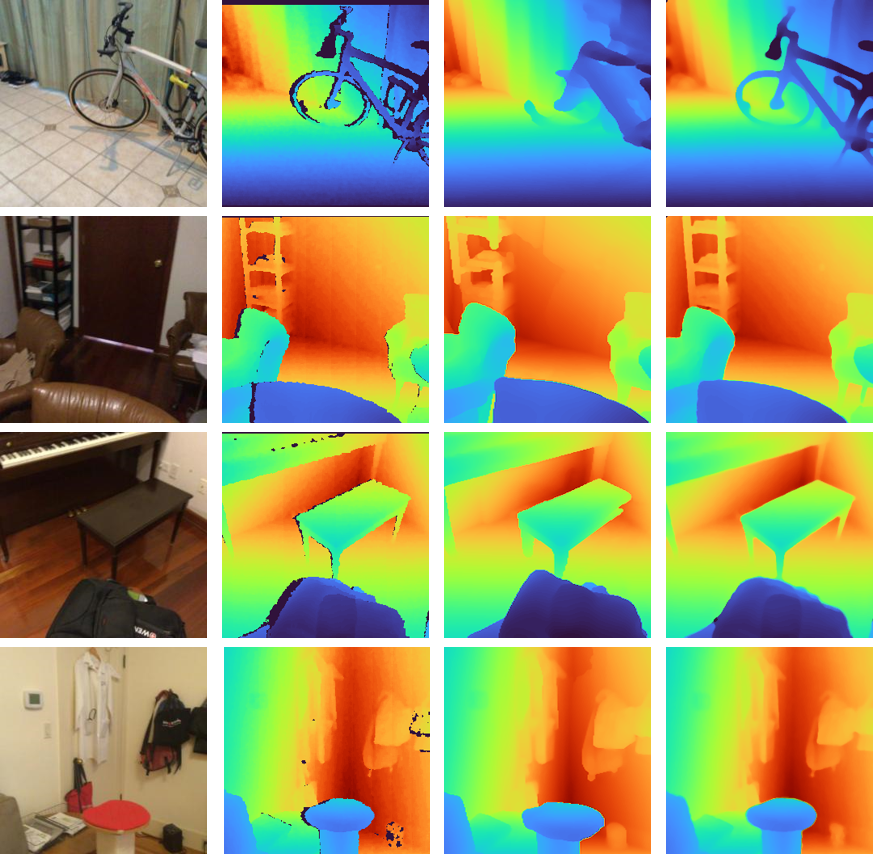}
    \end{center}
    \caption{Qualitative GT generation results of ScanNet \cite{dai2017scannet} and the proposed method. First column: RGB images; second column: input depth images; third column: GT generated based on \cite{dai2017scannet}; last column: GT generated using the proposed method.}
    \label{fig:gt_comparison}
\end{figure}

\subsubsection{Depth Rendering}
When the point clouds in the frame set are aligned to the local target frame, the GT depth image for the input target depth image is rendered by the averaged projection of the merged point cloud onto a rasterized image coordinate \cite{ravi2020pytorch3d}. Let $\textbf{d}_{i,k}\in [\textbf{d}_{i,1},\textbf{d}_{i,m}]$ be the $k^{th}$ nearest projected point to pixel $i$ when the 3D point $\textbf{p}_{i,k}$ is projected onto the rasterized image plane. Then, the refined depth value in pixel $i$ is formed as the weighted average of the nearest $m$ projected points. The exponential weight for $\textbf{d}_{i,k}$ is defined as $\omega_{i,k}=e^{-\widehat{\textbf{d}}_{i,k}}$, where $\widehat{\textbf{d}}_{i,k}={\textbf{d}_{i,k}}/{R^2}$ with radius $R$. Let depth value of point $\textbf{d}_{i,k}=z_{i,k}$; then, the weighted summation of $m$ depth values for pixel $i$ is computed as follows:

\begin{equation}
    \begin{split}
        \bar{\textbf{d}}_i=\sum_{k=1}^m{\widehat{\omega}_{i,k} \cdot z_{i,k}},\\
    \end{split}
    \label{eq:5}
\end{equation}
where $\bar{\textbf{d}}_i$ is the refined depth value for pixel $i$ and  $\widehat{\textbf{d}}_{i,k}$ is normalized weight factor. The closer the depth is, the more it is weighted to compute the refined depth value for the rasterized pixel. The generated image has reduced noise with the averaging manner, due to the averaging method, and the missing values are covered in the detected region in the neighbor frames.
\par

\begin{table}[ht]
\centering
\caption{Comparison of geometric structure between GT depth and original depth.}
\label{tab:table1}
\centering
\begin{tabular}{P{0.2\linewidth}|P{0.3\linewidth}P{0.3\linewidth}}
\hline
Method  & SSIM & $\mathcal{L}_S$ \\
\hline
ScanNet \cite{dai2017scannet} & 0.9314 &  2.3547 \\
\textbf{Ours} & \textbf{0.9709} & \textbf{0.1619} \\
\hline
\end{tabular}
\end{table}

\begin{figure*}[t]
    \begin{center}
    \includegraphics[width=\linewidth]{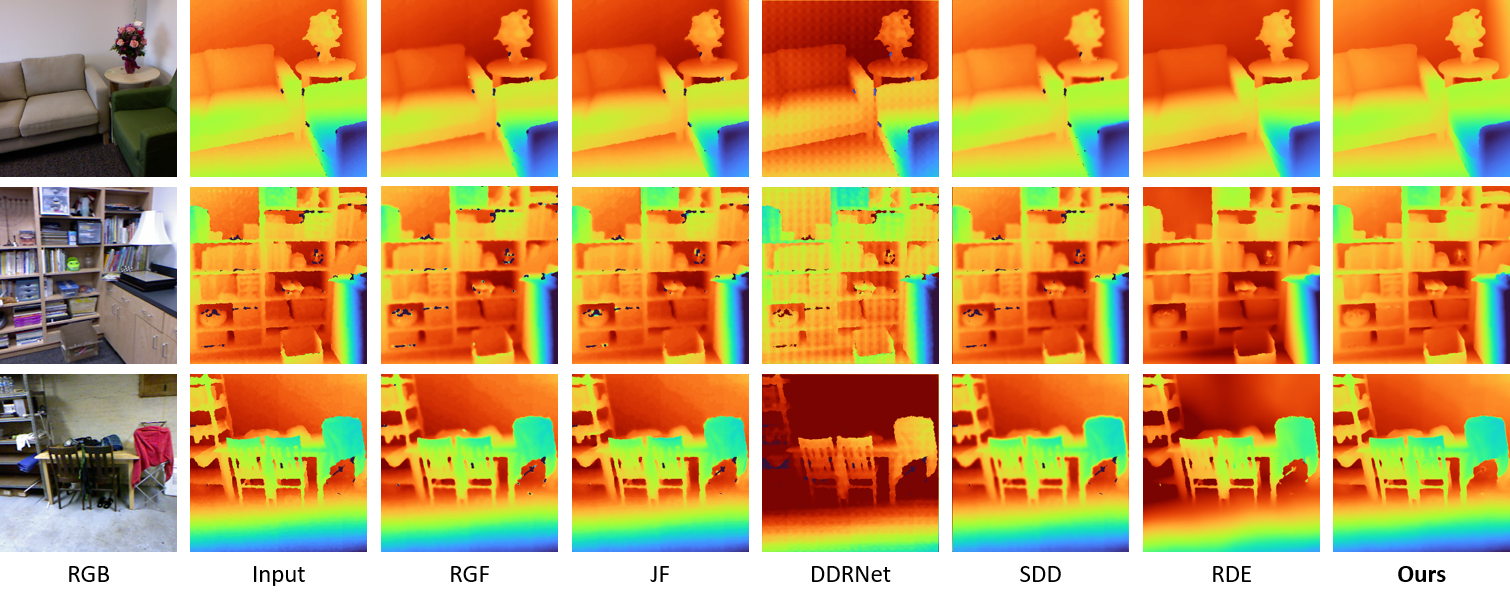}
    \end{center}
    \caption{Qualitative results of NYU-V2 \cite{silberman2012indoor} for the real-world dataset. In each row, the first and second columns show the original RGB and input depth images, respectively. The results of comparative methods are shown in columns three to seven (filter-based methods: third to fourth columns; deep-learning-based methods: fifth to seventh columns). The results from our dataset is shown in the last column.}
    \label{fig:nyu_qual}
\end{figure*}

\begin{figure*}[t!]
    \begin{center}
    \includegraphics[width=\linewidth]{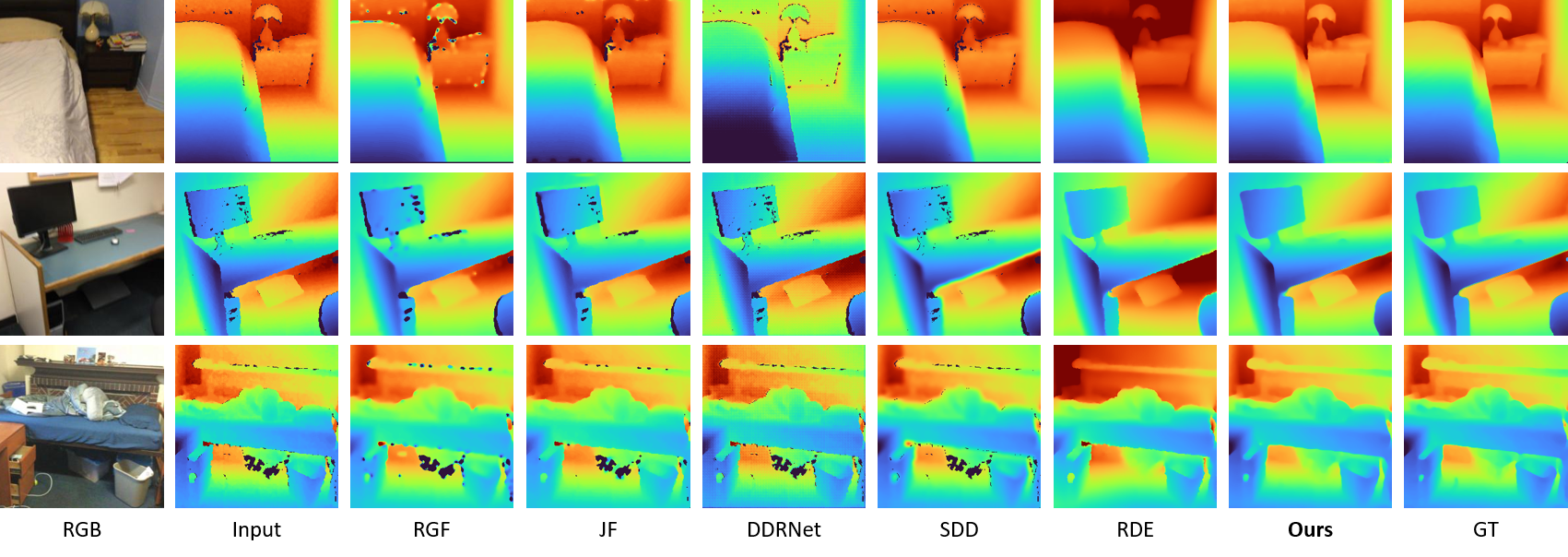}
    \end{center}
    \caption{Qualitative results of ScanNet \cite{dai2017scannet} for the real-world dataset. Similar to Fig. \ref{fig:nyu_qual}, the columns show the original RGB images, input depth images, and the results of the different methods. The GT depth images obtained using the proposed method are added in the last column for a visual comparison.}
    \label{fig:Scan_qual}
\end{figure*}

\section{Experimental Results}
\subsection{Materials}
A large-scale RGB-D dataset was required for the GT depth generation framework. We used the ScanNet \cite{dai2017scannet} dataset, which provides millions of RGB-D images of various indoor scenes. In our experiments, we defined the neighboring frames as three previous and successive frames each (i.e., a total of six frames) of the local target frame with a two-frame interval; thus, 3,000 input GT depth pairs were generated.
\par

We used the generated dataset for supervision in the depth enhancement framework to verify the superiority of our dataset. Note that the input RGB-D images were resized to 256$\times$256 pixels, and depth patches of 128$\times$128 pixels were randomly cropped six times from the depth pairs for training samples. Subsequently, we selected patch pairs based on the ratio of missing values that was less than 5$\%$ of the original patch area. Further, 15,787 input GT depth pairs were obtained, which consisted of 14,471 pairs for training and 1,316 pairs for validation. For a fair comparison of qualitative analysis, we used the depth images from the ScanNet \cite{dai2017scannet} and NYU-V2 \cite{silberman2012indoor} dataset as the real-world dataset, whereas those from SceneNet \cite{mccormac2017scenenet} were considered for the synthetic dataset. The comparison of quantitative analysis was evaluated using both the realistic and synthetic datasets, which were composed of 500 samples from ScanNet \cite{dai2017scannet} for the realistic dataset and 300 samples from SceneNet \cite{mccormac2017scenenet} for the synthetic dataset.
\par

\begin{figure*}[t]
    \begin{center}
    \includegraphics[width=\linewidth]{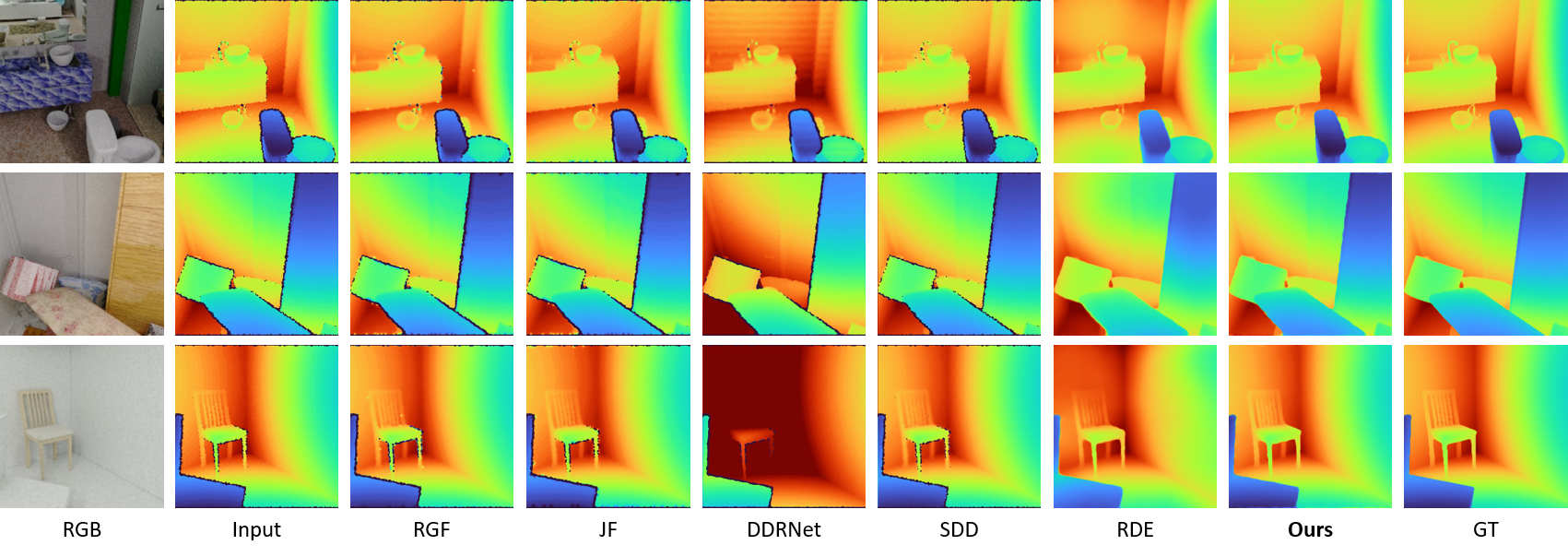}
    \end{center}
    \caption{Qualitative results of SceneNet \cite{mccormac2017scenenet} for the synthetic dataset. In each row, the first and last columns show the provided synthetic RGB and GT depth images, respectively. Noise-simulated RGB images using the method described in \cite{barron2013intrinsic} are illustrated in the second column. The remaining columns (columns 3-8) show depth enhancement results of the comparative methods and the proposed method.}
    \label{fig:scene_qual}
\end{figure*}

\subsection{Ground-truth (GT) Depth Generation}
 Figure \ref{fig:our_gt} illustrates the results of the generated GT depth images. The generated depth data preserved the geometric structures of the original depth image without misalignment errors. Figure \ref{fig:gt_comparison} shows qualitative comparison results between the proposed GT generation method and the previous ScanNet \cite{jeon2018reconstruction} method, which was constructed by global optimization of the pose parameters. The dataset constructed in \cite{jeon2018reconstruction} induced a misalignment error in certain local frames. Moreover, the reconstructed meshes also contained an over-smoothing problem, particularly in object boundaries, which can severely mislead the training of the original geometric structures. In contrast, the proposed method did not use redundant depth frames for the GT depth from a certain view frame. The estimated registration parameters in this study were optimized in an independent local frame set to alleviate the misalignment error. Subsequently, the generated GT depth image was directly rendered with the inverse-projected point cloud, according to (\ref{eq:5}), to mitigate the over-smoothing problem from the reconstructed meshes. 
 \par
 The preservation ability of the original geometric structures was evaluated to verify the superiority of the proposed GT dataset. We compared the structural similarity (SSIM) between the generated GT depth and original input depth, as proposed in \cite{jeon2018reconstruction, gu2020coupled}, and the maximum gradient magnitude distance in (\ref{eq:6}) (i.e., $\mathcal{L}_S$), which was defined for structure-preserving loss. From the results listed in Table \ref{tab:table1}, we can observe that our GT depth dataset outperforms the reconstruction-based method \cite{dai2017scannet} in both metrics. The superiority of our dataset is twofold: 1) the pose parameters optimized in the local frame set overcame the data misalignment error from the globally estimated parameters and 2) the over-smoothed mesh reconstruction error was alleviated by direct rendering of the merged point clouds.
 \par

\begin{table*}[ht]
\centering
\caption{Quantitative results on real-world dataset.}
\label{tab:table2}
\centering
\begin{tabular}{P{0.15\linewidth}|P{0.2\linewidth}P{0.2\linewidth}P{0.2\linewidth}}
\hline
Method  & SSIM & RMSE & MAE \\
\hline
RGF \cite{zhang2014rolling} & 0.8975 & 0.3734 & 0.1746 \\
JF \cite{shen2015mutual} & 0.8861 & 0.3518 & 0.1769 \\
DDRNet \cite{yan2018ddrnet} & 0.8605 & 0.4561 & 0.3824 \\
SDD \cite{sterzentsenko2019self} & 0.9296 & 0.3469 & 0.2355 \\
RDE \cite{jeon2018reconstruction} & 0.9267 & 0.2559 & 0.1831 \\
\textbf{Ours} & \textbf{0.9698} & \textbf{0.2258} & \textbf{0.1665}\\
\hline
\end{tabular}
\end{table*}

\begin{table*}[ht]
\centering
\caption{Quantitative results on synthetic dataset.}
\label{tab:table3}
\centering
\begin{tabular}{P{0.15\linewidth}|P{0.2\linewidth}P{0.2\linewidth}P{0.2\linewidth}}
\hline
Method  & SSIM & RMSE & MAE \\
\hline
RGF \cite{zhang2014rolling} & 0.9082 & 0.4251 & 0.2745 \\
JF \cite{shen2015mutual} & 0.9075 & 0.4269 & 0.2801 \\
DDRNet \cite{yan2018ddrnet} & 0.8179 & 0.5159 & 0.4413 \\
SDD \cite{sterzentsenko2019self} & 0.9047 & 0.3986 & 0.3624 \\
RDE \cite{jeon2018reconstruction} & 0.9066 & 0.2765 & 0.2019 \\
\textbf{Ours} & \textbf{0.9372} & \textbf{0.2648} & \textbf{0.1828}\\
\hline
\end{tabular}
\end{table*}

\subsection{Depth Enhancement}
We compared the depth enhancement results based on our GT depth dataset against the results obtained using both the traditional filter-based and deep-learning-based approaches using the real-world datasets; rolling guidance filtering (RGF \cite{zhang2014rolling}) and joint filtering (JF \cite{shen2015mutual}) were considered for the filter-based methods, whereas reconstruction-based depth enhancement (RDE \cite{jeon2018reconstruction}), depth denoising and refinement network (DDRNet \cite{yan2018ddrnet}), and self-supervised depth denoising (SDD \cite{sterzentsenko2019self}) were considered for the learning-based-methods. Note that the refinement part of the network in the DDRNet have been omitted for a fair comparison as in \cite{sterzentsenko2019self}. The depth enhancement results of the comparative methods were evaluated using both real-world and synthetic depth datasets in qualitative and quantitative manners (Figs. \ref{fig:nyu_qual}, \ref{fig:Scan_qual}, and \ref{fig:scene_qual}). The quantitative comparison was evaluated based on the SSIM, root mean square error (RMSE), and mean absolute error (MAE) metrics. The default parameters of the filter-based methods (RGF \cite{zhang2014rolling}, JF \cite{shen2015mutual}) were used, as in their provided codes.
\par

\subsubsection{Learning Architecture}
A deep Laplacian pyramid depth image enhancement network (LapDEN) was adopted \cite{jeon2018reconstruction} to train the deep neural network for the depth enhancement framework using the constructed GT depth dataset. The network predicts an enhanced depth image from a coarse to fine scale using a progressive upsampling scheme in an image pyramid without loss of scale-variant features based on the deep Laplacian pyramid network architecture \cite{lai2017deep}. Please refer to the original manuscript for more details \cite{jeon2018reconstruction}.
\par

\subsubsection{Loss Functions}
Given the original input depth image $\textbf{x}$ and the corresponding GT depth image $\textbf{y}$, the loss functions to estimate the enhanced depth image $\hat{\textbf{y}}$ are defined as $\mathcal{L}=\mathcal{L}_D(\hat{\textbf{y}},\textbf{y}) + \lambda_s\mathcal{L}_S(\hat{\textbf{y}},\textbf{x})$ \cite{jeon2018reconstruction}, where $\mathcal{L}_D$ and $\mathcal{L}_S$ indicate data loss and structure preserving loss, respectively. The $\mathcal{L}_D$ is a combination of $L_1$ distances between $\hat{\textbf{y}}$ and $\textbf{y}$ in terms of depth, depth gradient, and surface normal. $\mathcal{L}_D$ was adopted in this study to directly train the network using the GT depth geometry for enhanced depth prediction. The other structure-preserving loss term $\mathcal{L}_S$ is defined in the original paper \cite{jeon2018reconstruction} as follows:
\begin{equation}
       \mathcal{L}_S=\frac{1}{N}\sum_p{\left( \max_{q\in\Omega(p)}|\nabla\hat{y}_q|-\max_{q\in\Omega(p)}|\nabla x_q| \right)}^2,
       \label{eq:6}
\end{equation}
where $N$ is the total number of pixels and $\Omega(p)$ denotes a local window centered at pixel $p$. $\mathcal{L}_S$ was proposed to calculate similarity between the geometric structures of $\hat{\textbf{y}}$ and $\textbf{x}$ by imposing the maximum gradient magnitude loss around edge pixels. The authors utilized the original input depth $\textbf{x}$ for supervision instead of GT depth $\textbf{y}$ to prevent the data misalignment errors between the input and the GT depth image. However, the maximum gradient value around the heavy noise and the missing values, which may have contained in the input depth $\textbf{x}$, can disturb the training of the depth geometry obtained from the GT dataset. In this study, our GT depth data were used as supervision for the $\mathcal{L}_S$ term rather than the raw input depth data. That is, the loss functions from the original paper were modified as $\mathcal{L}=\mathcal{L}_D(\hat{\textbf{y}},\textbf{y}) + \lambda_s\mathcal{L}_S(\hat{\textbf{y}},\textbf{y})$, where $\nabla x_q $ is substituted by $\nabla y_q$. Owing to the pairwise depth dataset constructed using our proposed framework, our method does not suffer from the data misalignment problem. The precise GT depth dataset trains a more elaborate geometric structure from the GT depth dataset.
\par

\subsubsection{Real-world Dataset Evaluation}
Figure \ref{fig:nyu_qual} and \ref{fig:Scan_qual} present the qualitative analysis results of depth enhancement based on the NYU-V2 \cite{silberman2012indoor} and ScanNet \cite{dai2017scannet} dataset, respectively. Because DDRNet \cite{yan2018ddrnet} and SDD \cite{sterzentsenko2019self} are only utilized for depth denoising, they cannot cope with the missing depth values appropriately. Although filter-based methods covered the missing holes marginally, they were inadequate when retrieving the entire scene. Only RDE \cite{jeon2018reconstruction} performed with promising results in the comparative methods; however, partial noisy regions remained, especially in object boundaries, which originated from the data misalignment error between the GT depth and original depth data. Furthermore, the method failed to recover thin objects in certain cases (e.g., first row of the sixth column in Fig. \ref{fig:Scan_qual}) owing to the over-smoothed mesh reconstruction error in its dataset. Table \ref{tab:table2} presents a quantitative comparison of the methods. The results demonstrate that the depth enhancement results from our dataset outperformed those of the other state-of-the-art comparative methods. Owing to the superiority of our GT dataset, as shown in Fig. \ref{fig:gt_comparison} and Table \ref{tab:table1}, the proposed GT dataset were used as benchmark dataset for the evaluations rather than the previously benchmarked dataset \cite{dai2017scannet}.
\par

\subsubsection{Synthetic Dataset Evaluation}
Synthetic depth data were also evaluated to clarify the superiority of the proposed GT dataset. Figure \ref{fig:scene_qual} illustrates the qualitative results of the depth enhancement for the synthetic RGB-D dataset provided by SceneNet \cite{mccormac2017scenenet}. To simulate the raw input depth images for the evaluation, Kinect-style noise \cite{barron2013intrinsic} was added to the original synthetic GT depth images (i.e., second column: simulated input depth images, last columns: original synthetic GT images in Fig. \ref{fig:scene_qual}). The RDE \cite{jeon2018reconstruction} achieved promising results similar to the real-world case; however, inferior results were observed when recovering thin objects (e.g., first row of seventh column in Fig. \ref{fig:scene_qual}). Conversely, the results from our dataset show the successful recovery of such structures when compared to the other methods. The comparative results of quantitative analysis are listed in Table \ref{tab:table3}. The results demonstrated that our proposed depth enhancement outperformed the other state-of-the-art methods on the synthetic dataset case also.
\par

\subsection{Implementation Details}
\subsubsection{GT Depth Generation}
The pose parameters in the local frame set (i.e., target frame and six neighboring frames) were overfit trained based on a single batch and a $10^{-4}$ learning rate for 30 epochs without weight decay for the frame set registration step. The given intrinsic matrix of the camera was modified according to the input image size (i.e., 256$\times$256). The remaining parameters are identical to those suggested in \cite{el2021unsupervisedr}. 
\par

\subsubsection{Depth Image Enhancement}
The LapDEN network was trained using our dataset with a three-level spatial resolution for the learning stage, based on same parameters presented in \cite{jeon2018reconstruction}. We only have adjusted the coefficient $\lambda_s$ for the structure-preserving loss term $\mathcal{L}_S$ in (\ref{eq:6}) from 10 to 5, because the GT depth in both loss terms can successfully preserve the original structure when compared to the original paper.
\par

\section{Discussions}

Accurate 3D perception using depth-sensing devices is a prerequisite for many computer vision applications. However, the inherent noise in the single-view environment of most commercial depth cameras (e.g., distance, light source, and occlusion) is severe in downstreaming tasks. Recent research has proposed deep-learning-based approaches for the single-view depth enhancement of the data obtained from depth cameras, which typically train the networks using a high-quality GT depth dataset. Because the performance of deep-learning-based methods is primarily dependent on the quality of the GT dataset, the construction of a high-quality depth dataset is essential.
\par

Inspired by the fact that most frames in local spatial regions overlap considerably, our method leverages multiple independent neighboring frames for the generated high-quality GT depth dataset. Our method proposes a multiview-based dataset generation method using a local frame set. When compared to the previous approaches, the proposed method significantly reduced misalignment errors based on an unsupervised metric. The major difference from the previous approaches is that our training units were based on a local frame set rather than global frames. Although the single-view depth from the depth camera contains inherent noise, the proposed method enables the construction of a reliable GT dataset using a pure RGB-D stream dataset without any other supervision. The dataset can increase the performance of most real-world GT-based depth enhancement tasks based on the proposed high-quality supervision; moreover, our method introduces a new benchmarking standard for the performance evaluation metrics. Further, the dataset generation pipeline can be combined with several other deep-learning-based 3D computer vision applications as a fundamental process in high-precision depth acquisition.
\par

\section{Conclusions}
In this study, we presented a high-quality GT depth dataset generation method based on an unsupervised training scheme using an unlabeled RGB-D stream dataset. The GT depth image was obtained using an overfit-trained unsupervised point cloud registration scheme. The proposed method successfully generated a high-quality depth dataset that preserved the geometric structure of the original depth. The experimental results demonstrated that the generated dataset was superior to benchmarked dataset, and the depth enhancement results showed superior performance when compared to other state-of-the-art depth enhancement methods. The major contribution of our study was to obtain an accurate GT depth dataset. This dataset can be used as a supervision dataset for downstreaming tasks and provides a new benchmarking standard for performance evaluations in various 3D computer vision applications.
\par

\ifCLASSOPTIONcaptionsoff
  \newpage
\fi

\bibliographystyle{IEEEtran} 
\bibliography{IEEEabrv, references}

\end{document}